\documentclass[11pt,a4paper]{article}
\usepackage[hyperref]{acl2020}
\usepackage{times}
\usepackage{latexsym}
\usepackage{times}
\usepackage{epsfig}
\usepackage{graphicx}
\usepackage{amsmath}
\usepackage{amssymb}
\usepackage{amsfonts}
\usepackage{times}
\usepackage{url}
\usepackage{bm}
\usepackage{latexsym}
\usepackage{graphicx}
\usepackage{makecell}
\usepackage{subcaption}

\newcommand{\ie}{\textit{i}.\textit{e}., }
\newcommand{\eg}{\textit{e}.\textit{g}. }

\usepackage{microtype}

\aclfinalcopy 
\title{Unsupervised Multimodal Neural Machine Translation with \\Pseudo Visual Pivoting}

\author{Po-Yao Huang$^1$, Junjie Hu$^1$, Xiaojun Chang$^2$, Alexander Hauptmann$^1$ \\
  $^1$Language Technologies Institute, Carnegie Mellon University \\
  $^2$Faculty of Information Technology, Monash University \\
  \texttt{\{poyaoh, junjieh, alex\}@cs.cmu.edu, cxj273@gmail.com}}
\date{}

\begin{document}
\maketitle
\begin{abstract}
Unsupervised machine translation (MT) has recently achieved impressive results with monolingual corpora only. 
However, it is still challenging to associate source-target sentences in the latent space.
As people speak different languages biologically share similar visual systems, the potential of achieving better alignment through visual content is promising yet under-explored in unsupervised multimodal MT (MMT).
In this paper, we investigate how to utilize visual content for disambiguation and promoting latent space alignment in unsupervised MMT.
Our model employs multimodal back-translation and features pseudo visual pivoting in which we learn a shared multilingual visual-semantic embedding space and incorporate visually-pivoted captioning as additional weak supervision.
The experimental results on the widely used Multi30K dataset show that the proposed model significantly improves over the state-of-the-art methods and generalizes well when the images are not available at the testing time.
\end{abstract}


\section{Introduction}
Neural machine translation~\cite{KalchbrennerB13,SutskeverVL14} has achieved near human-level performance~\cite{wu2016}.
However, its effectiveness strongly relies on the availability of large-scale parallel corpora. 
Unfortunately, preparing the parallel data remains a challenge as there are more than 6,500 languages in the world, and recruiting translators with bilingual or multilingual knowledge to cover all those languages is impractical.

As a result, developing methods alleviating the need of well-annotated large parallel corpora has recently attracted increasing attention in the community. 
These methods fall into two broad categories. The first type of methods use a third language as the pivot~\cite{firat2016,chen2017,cheng2017,johnson17} to enable zero-resource translation. 
Although the progress is encouraging, pivoting with a third language still demands bilingual knowledge for collecting large-scale parallel source-pivot and pivot-target corpora.
The second type of methods explore unsupervised approaches~\cite{conneau2017word,Artetxe2018,lample2017unsupervised}
have recently achieved impressive translation quality.
These methods rely only on monolingual data and back-translation~\cite{sennrich2016improving}.
However, as discussed in~\cite{lample_2018_phrase}, the alignment of source-target sentences is uncertain and highly subject to proper initialization.

Using visual content for unsupervised MT~\cite{chen2018zero,su2019} is a promising solution for pivoting and alignment based on its availability and feasibility.
Abundant multimodal content in various languages are available online (\eg Instagram and YouTube). 
It is also easier to recruit monolingual annotators to describe an image than to find multilingual translators to translate sentences.
Importantly, visual content is eligible to improve the alignments in the language latent spaces since the physical visual perception is similar among people speaking different languages (\eg similar ``blue car'' for a German and a French).

Based on these insights, we propose a novel unsupervised multimodal MT framework incorporating images as pseudo pivots promoting latent space alignment. 
In addition to use features of visual objects for multimodal back-translation,
we align a shared multilingual visual-semantic embedding (VSE) space via leveraging disjoint image-sentence pairs in different languages. 
As illustrated in Figure~\ref{fig_pivot}, for sentences approximately pivoted by similar images (src-\textit{img}-tgt), 
drawing embeddings of corresponding image-sentence pairs closer results in better alignments of semantically equivalent sentences in the language latent spaces.
Inspired by back-translation, we further explore another pseudo pivoting strategy, which approximates multilingual sentence pairs (\textit{src}-img-\textit{tgt}) conditioned on a real image via captioning.
Instead of using annotation of images for pivoting as in~\cite{chen2018zero},
we generate sentences in two languages pivoted on the real image, and then approximately pairing them as weak supervision for training unsupervised MT system.
This approach is analogous to a cross-modal version of back-translation.

We make the following contributions: 
(1) Building a unified view of employing visual content for pseudo pivoting.
(2) We learn and improve the alignments in the shared multilingual multimodal embedding space for unsupervised MMT with disjoint image-text pairs in different languages.
(3) Our model achieves state of the art on Multi30K and generalizes well to the text-only scenario.

\section{Background}

\noindent\textbf{Neural Machine Translation}
Typical NMT models are based on the encoder-decoder framework with attention~\cite{attn_nmt}.
Let $\mathbf{x}=(x_1, \cdots, x_N)$ denotes a source sentence and $\mathbf{y}=(y_1, \cdots, y_M)$ denotes a target sentence, where $(\mathbf{x},\mathbf{y}) \in (\mathcal{X},\mathcal{Y})$. 
The encoder-decoder model learns to estimate the following likelihood from the source sentence to the target sentence:
\begin{equation}\label{mt_likelihood}
p_{x\rightarrow y}(\mathbf{y}|\mathbf{x})=\prod_{i=1}^M p(y_i|\mathbf{y}_{<i},\mathbf{x})
\end{equation}

When a parallel corpus is available, the maximum likelihood estimation (MLE) is usually adopted to optimize the (source to target language) NMT model by minimizing the following loss:
\begin{equation}\label{mt_loss}
\mathcal{L}_{x \rightarrow y}^{MT}= \mathbb{E}_{(\mathbf{x},\mathbf{y}) \sim (\mathcal{X}, \mathcal{Y})}\left[ -\text{log } p_{x\rightarrow y}(\mathbf{y}|\mathbf{x}) \right] 
\end{equation}

Among all encoder-decoder models, the Transformer~\cite{vaswani2017attention} architecture recently achieves state-of-the-art translation quality. 
Instead of using recurrent or convolutional operations, it facilitates multi-head self-attention~\cite{selfatt}. 
In this paper, we choose the Transformer as the underlying architecture for both the translation and the captioning modules.

\noindent\textbf{Unsupervised Machine Translation}
While conventional MT systems rely on the availability of a large parallel corpus,
translation with zero-resource (unsupervised MT)~\cite{lample2017unsupervised,Artetxe2018,lample_2018_phrase} has drawn increasing research attention.
Only monolingual sentences are presented at the training and validation phase, \ie only $\mathbf{x} \in \mathcal{X}$ and $\mathbf{y} \in \mathcal{Y}$ are available.

Successful unsupervised MT systems share several common principles. 
First, they require the pre-training step to initialize the model and establish strong monolingual language models properly. 
For example, XLM~\cite{xlm} utilizes the masked language model objective in BERT~\cite{bert}. 
MASS~\cite{mass} utilizes a span-based sequence-to-sequence masking objective for language model pre-training.

Second, these systems transform the unsupervised problem into a weakly or self-supervised one by automatically generating pseudo sentence pairs via back-translation~\cite{sennrich2016improving}. 
The idea behind can be analogous to the cycle-consistency objective in CycleGAN~\cite{cyc} for image-image translation with unpaired data.
Specifically, let us denote by $h^*(\mathbf{y}) = (\hat{x}_1, \cdots, \hat{x}_N)$
the sentence in the source language inferred from $\mathbf{y}\in \mathcal{Y}$
such that $h^*(\mathbf{y})= \text{argmax }p_{y \rightarrow x}(\mathbf{x}|\mathbf{y})$.
Similarly, let us denote by $g^*(\mathbf{x}) = (\hat{y}_1, \cdots, \hat{y}_M)$ the sentence in the target language inferred from $\mathbf{x}\in \mathcal{X}$ such that $g^*(\mathbf{x})= \text{argmax } p_{x \rightarrow y}(\mathbf{y}| \mathbf{x})$.
Then the ``pseudo'' parallel sentences $\left(h^*(\mathbf{y}), \mathbf{y} \right)$ and $\left(\mathbf{x}, g^*(\mathbf{x})\right)$ can be further used to train two two MT models ($\mathcal{X}\rightarrow\mathcal{Y}$ and $\mathcal{Y}\rightarrow\mathcal{X}$) by minimizing the following back-translation loss:
\begin{equation}
\begin{aligned}\label{umt_loss}
\mathcal{L}_{x \leftrightarrow y}^{BT} &= \mathbb{E}_{\mathbf{x} \sim \mathcal{X}}\left[ -\text{log } p_{y\rightarrow x}(\mathbf{x}|g^*(\mathbf{x})) \right] \\
&+\mathbb{E}_{\mathbf{y} \sim \mathcal{Y}}\left[ -\text{log } p_{x\rightarrow y}(\mathbf{y}|h^*(\mathbf{y})) \right]
\end{aligned}
\end{equation}
 
Although reinforcement learning-based approaches~\cite{dual} and Gumbel-softmax reparametrization~\cite{Gumbel} have been used to handle back-propagation thorough non-differentiable ``argmax'' predictions.
in this paper, we do not back-propagate through $h^*(\mathbf{y})$ and $g^*(\mathbf{x})$ to simplify the training process.

\begin{figure*}[th!]
    \centering
    \includegraphics[width=\linewidth]{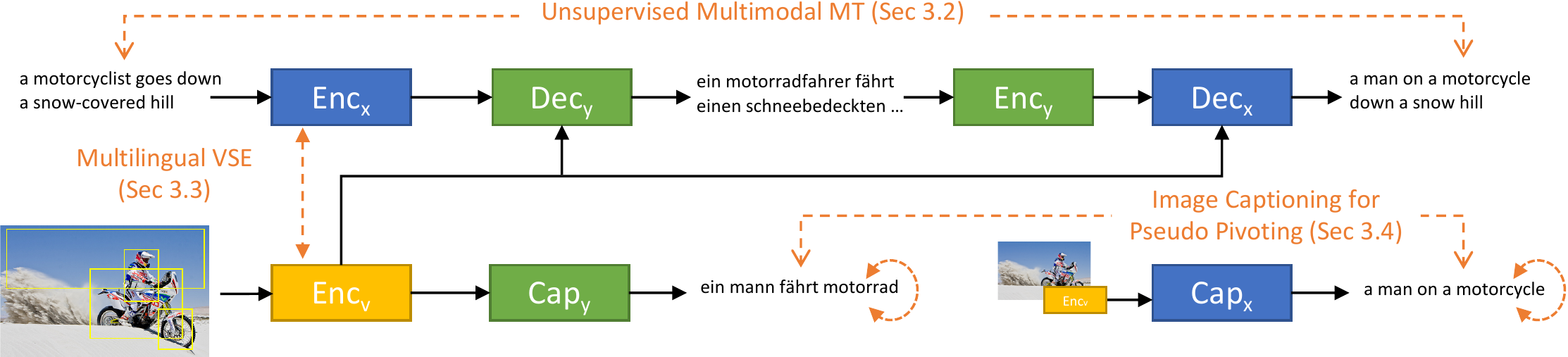}
    \caption{The proposed model structure (English$\leftrightarrow$German). We incorporate visual objects for unsupervised multimodal MT and improve the language latent space alignment with \textit{pseudo visual pivoting}  (\S\ref{sec_vse}-\S\ref{sec_cap}).}
    \label{sys}
\end{figure*}

\section{Unsupervised Multimodal Machine Translation}
As illustrated in Figure~\ref{sys}, our model is composed of seven modules: Two encoder-decoder pairs for translation, two decoders for captioning, and one shared visual encoder. 
In this section, we first detail our basic MMT model architecture and the unsupervised setup. Then we introduce pseudo visual pivoting: learning multilingual VSE and pivoted captioning.

\subsection{Multimodal MT}
Multimodal machine translation~\cite{mmt_task} (MMT) considers additional images as a complementary information source for MT. 
An image $\mathbf{z}$ and the description in two languages form a triplet $(\mathbf{x}, \mathbf{y},\mathbf{z}) \in (\mathcal{X},\mathcal{Y}, \mathcal{Z})$. 
The Transformer encoder reads the source sentence and encodes it with hierarchical self-attention into $\mathbf{h}^x=\{\mathbf{h}_1^x,\cdots,\mathbf{h}_N^x\}, \mathbf{h}_i^x \in \mathbb{R}^d$, where $d$ is the dimension of the embedding space. 
The visual encoder encodes the image into $\mathbf{h}^z=\{\mathbf{h}_1^z,\cdots,\mathbf{h}_K^z\}, \mathbf{h}_i^z \in \mathbb{R}^d, K_{\text{max}}=36$. Most previous work~\cite{chen2018zero,su2019} use 2D ($K=14\times14$) feature maps of ImageNet pre-trained ResNet~\cite{he2016deep}. 
In contrast, we utilize the regional features of $K$ salient visual objects in an image extracted by Faster-RCNN~\cite{ren2015faster} 
and a 1-layer MLP as the encoder to encode visual objects.


Various attention strategies for sequence-to-sequence learning have been addressed in~\cite{attns}. 
Our model employs the hierarchical multi-head multimodal attention for decoding.
For decoding at time stamp $i$, the textual attention $\text{Attn}(\mathbf{h}_i^y, \mathbf{h}^x)$ computes the context vector $\mathbf{c}_i = \sum_j \alpha_j \mathbf{h}_j^x$ via a attention-based alignment $\alpha_j=\text{Align}(\mathbf{h}_i^y$, $\mathbf{h}_j^x)$, where $\sum_j\alpha_j=1$ and $\mathbf{h}_i^y$ is the decoder state.
Essentially, the one-head attention in Transformer is implemented as $\mathbf{c}_i=\text{softmax}(\mathbf{Q}_i(\mathbf{K}^x)^\top/\sqrt{d})\mathbf{V}^x$ where $\{ \mathbf{Q}, \mathbf{K}^x, \mathbf{V}^x\}$ are the packed $d$-dimensional \textit{Query, Key, Value} vectors, which are the mapped and packed version of $\{\mathbf{h}_i^y, \mathbf{h}^x, \mathbf{h}^x\}$.
For decoding with encoded visual and textual inputs, we utilize multimodal attention to compute the context vector $\mathbf{c}_i$:
\begin{equation}
    \mathbf{c}_i^x = \text{Attn}(\mathbf{h}^y_{i-1},\mathbf{h}^x) + \lambda_v \text{Attn}(\mathbf{h}^y_{i-1}, \mathbf{h}^z)
\end{equation}
In practice we set $\lambda_v=1.0$. Our multimodal decoder models the likelihood to predict the next token as:
\begin{equation}\label{mmt_likelihood}
p(y_i|\mathbf{y}_{<i},\mathbf{x},\mathbf{z})= \text{softmax}(f(\mathbf{c}_i, y_{i-1}, \mathbf{h}^y_{i-1} ),
\end{equation}
where $f(.)$ denotes the aggregated non-linear feature mapping in Transformer.

\begin{figure*}
    \centering
    \includegraphics[width=\linewidth]{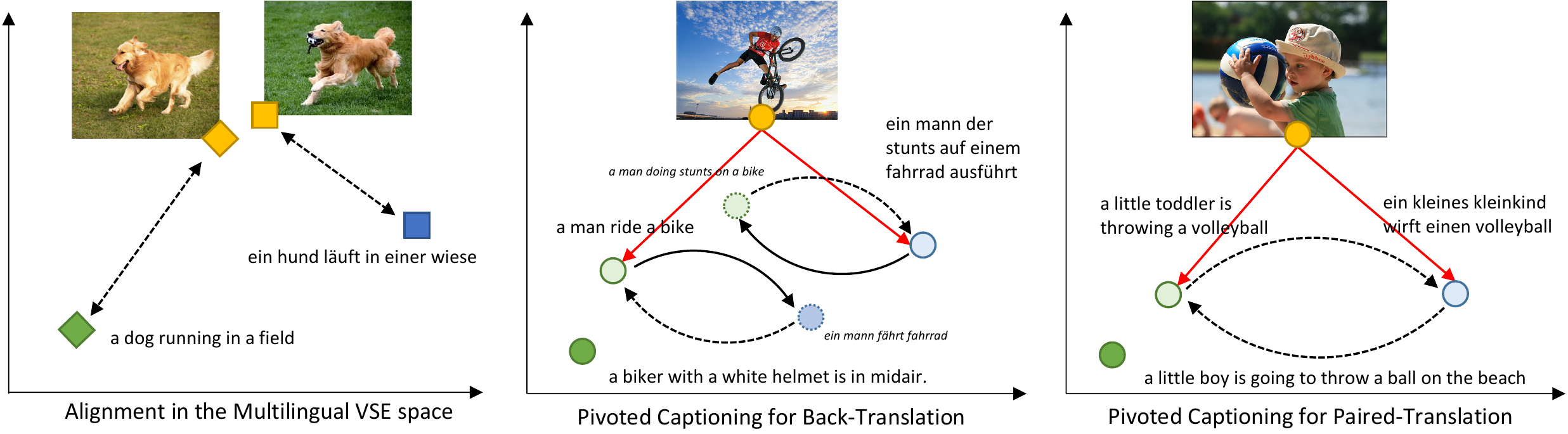}
    \caption{Pseudo visual pivoting: (1) multilingual VSE (src-\textit{img}-tgt, in fact src-img\textsubscript{1}, tgt-img\textsubscript{2}), and (2) pivoted captioning (\textit{src}-img-\textit{tgt}). The \textit{italic} items do not exist and are approximated (pseudo).
    (src, img, tgt) is colored in (green, yellow, blue).
    Solid red and black lines indicate captioning and translation without updates. Encoder-decoder are updated with dashed lines to improve the alignments in the multilingual multimodal embedding space.}
    \label{fig_pivot}
\vspace{-0.5em}
\end{figure*}

\subsection{Unsupervised Learning}
Unsupervised multimodal MT~\cite{nakayama2017zero,chen2018zero,su2019} poses a new yet challenging problem. On both the source and target sides, only non-overlapping monolingual multimodal data are presented for training and validation. 
Specifically, the data available are: $(\mathbf{x}, \mathbf{z}_x)\in (\mathcal{X}, \mathcal{Z})$, $(\mathbf{y}, \mathbf{z}_y)\in (\mathcal{Y}, \mathcal{Z})$, such that $\{ \mathbf{x} \} \cap \{ \mathbf{y} \}= \phi$, $\{ \mathbf{z}_x \} \cap \{ \mathbf{z}_y \} = \phi$. Note that there are no parallel translation pairs available (unsupervised), and the images are mutually exclusive for different languages.

For multimodal back-translation, the generated pseudo target sentence conditioned on the source sentence and image can be re-written as
$g^*(\mathbf{x}, \mathbf{z}_x) = \text{argmax } p_{xz \rightarrow y}(\mathbf{y}| \mathbf{x},\mathbf{z}_x)$, where $p_{xz\rightarrow y}(\mathbf{y}|\mathbf{x},\mathbf{z})=\prod_{i=1}^M p(y_i|\mathbf{y}_{<i},\mathbf{x}, \mathbf{z})$. 
Similar for $p_{yz\rightarrow x}(\mathbf{x}|\mathbf{y},\mathbf{z})$ and $h^*(\mathbf{y}, \mathbf{z}_y)$.
For unsupervised multimodal MT, the multimodal back-translation objective can be extended as:
\begin{equation}
\begin{aligned}
\mathcal{L}_{x\leftrightarrow y}^{MBT}&=\mathbb{E}_{(\mathbf{x},\mathbf{z}_x) }\Big[ \text{-log } p_{yz\rightarrow x}\left(\mathbf{x}|g^*(\mathbf{x},\mathbf{z}_x), \mathbf{z}_x\right) \Big] \\
&+\mathbb{E}_{(\mathbf{y},\mathbf{z}_y)}\Big[\text{-log } p_{xz\rightarrow y}\big(\mathbf{y}|h^*(\mathbf{y},\mathbf{z}_y),\mathbf{z}_y)\big) \Big]
\end{aligned}\label{ummt_loss}
\end{equation}
We simplify the notation of expectation for clarity.

Aligning the latent spaces of the source and target languages without supervision is challenging, as discussed in~\cite{lample_2018_phrase}.
However, as people speak different languages biologically share similar visual systems, we envision that the shared visual space can serve as the pivot for alignment.
Unlike most previous work~\cite{chen2018zero,su2019} treating images merely as a feature, we propose two visual pivoting approaches:
(1) Aligning the multilingual VSE space; (2) Image pseudo pivoting via captioning.
As illustrated in Figure~\ref{fig_pivot}, for (1), we use images as the approximate pivots connecting real non-parallel sentences. (src-\textit{img}-tgt.)
In (2), for each pivoting real image, we generate captions in both languages to construct ``pseudo'' source-target sentence pairs. (\textit{src}-img-\textit{tgt}), where the \textit{italic} item is ``pseudo''.
We collectively term the proposed approach \textit{pseudo visual pivoting}.

\subsection{Multilingual Visual-Semantic Embedding}\label{sec_vse}
We posit that for $\mathcal{X},\mathcal{Y},\mathcal{Z}$, the two language spaces $\mathcal{X},\mathcal{Y}$ could be properly associated by respectively aligning two monolingual VSE spaces $\mathcal{X}\leftrightarrow\mathcal{Z}$ and $\mathcal{Y}\leftrightarrow\mathcal{Z}$.
We leverage the contrastive objective in cross-modal retrieval~\cite{KirosSZ14,ann} for aligning multimodal inputs in the shared VSE space where the embeddings are close if they are semantically associated or paired.

Specifically, we generalize the fine-grained (object-level and token-level), monolingual textual-to-visual, and visual-to-textual attention~\cite{scan,oan} into the multilingual setup.
For fine-grained image-sentence alignment, let $s_{ij}=\text{cos}(\mathbf{h}_i^x, \mathbf{h}_j^z)$ denotes the cosine similarity between the $i$-th encoded token and the $j$-th encoded visual object.
The image-sentence similarity can be measured by averaging the cosine similarities between the visually-attend sentence embeddings and the visual embeddings of the objects.
The visually-attended sentence embeddings $\mathbf{h}^{zx}$ are the weighted combination of the encoded tokens $\mathbf{h}^x$.
Precisely, we compute $\mathbf{h}_j^{zx} = \sum_{i=1}^N \alpha_{ij} \mathbf{h}_i^x$, where $j=1\cdots K$ and $\alpha_{ij}=\text{softmax}_i(s_{ij})$. 
Let us denote by $S(\textbf{x},\textbf{z})=\frac{1}{2K}\sum_{j=1}^K \text{cos}(\mathbf{h}^{zx}_j, \mathbf{h}^{z}_j)+\frac{1}{2N}\sum_{i=1}^N \text{cos}(\mathbf{h}^{xz}_i, \mathbf{h}^{x}_i)$ as the image-sentence similarity, the contrastive triplet loss encouraging image-sentence alignment in the VSE space can be written as:
\begin{equation}\label{loss_vse}
\begin{aligned}
 \mathcal{L}_{c}(\mathbf{x},\mathbf{z}) &=
 \underset{\tilde{\mathbf{x}}}{\text{max}} \big[\gamma - S(\mathbf{x},\mathbf{z}) + S(\tilde{\mathbf{x}}, \mathbf{z})\big]_+ \\
 &+\underset{\tilde{\mathbf{z}}}{\text{max}}\big[\gamma - S(\mathbf{x},\mathbf{z}) + S(\mathbf{x},\tilde{\mathbf{z}})\big]_+,
\end{aligned}
\end{equation}
where $[.]_+$ is the hinge function, and $\tilde{\mathbf{x}}$ and $\tilde{\mathbf{z}}$ are the non-paired (negative) instances for $\mathbf{x}$ and $\mathbf{z}$. Intuitively, when the loss decreases, the matched images and sentences will be drawn closer down to a margin $\gamma$ than the hardest non-paired ones.
Formally, we minimizing the following objective for cross-modal alignments in the two VSE spaces:
\begin{equation}
\mathcal{L}_{x,y,z}^{VSE}=\mathbb{E}_{(\mathbf{x},\mathbf{z}_x)}\Big[ \mathcal{L}_{c}(\mathbf{x}, \mathbf{z}_x)\Big] + \mathbb{E}_{(\mathbf{y},\mathbf{z}_y)}\Big[ \mathcal{L}_{c}(\mathbf{y}, \mathbf{z}_y)\Big]
\end{equation}

\subsection{Image Captioning for Pseudo Pivoting}\label{sec_cap}
Inspired by back-translation with monolingual corpora, we propose a novel cross-modal approach to generate weakly-supervised pairs to guide language space alignment for unsupervised MMT.
Precisely, we leverage image captioning to synthesize pseudo sentence pairs (pivoted and conditioned on the image)  for back-translation and paired-translation.

\noindent\textbf{Image Captioning}
Image captioning models are akin to MT models besides the non-sequential visual encoder. 
For example, an image-to-source captioning model estimates the likelihood as 
$p_{z\rightarrow x}(\mathbf{x}|\mathbf{z})=\prod_{i=1}^N p(x_i|\mathbf{x}_{<i},\mathbf{z})$, where $\mathbf{z}$ is the encoded image. 
Essentially, the captioning model learns to minimize the following loss:
\begin{equation}\label{cap_loss}
\mathcal{L}_{z \rightarrow x}^{CAP}= \mathbb{E}_{(\mathbf{z}_x,\mathbf{x})}\left[ -\text{log } p_{z\rightarrow x}(\mathbf{x}|\mathbf{z}_x) \right] 
\end{equation}


As illustrated in Figure~\ref{fig_pivot}, we incorporate two captioning models $\mathcal{Z}\rightarrow\mathcal{X}$ and $\mathcal{Z}\rightarrow\mathcal{Y}$ to generate additional ``pseudo'' parallel sentences pivoted on the image as additional weak supervision to better align language latent spaces in unsupervised MMT.
For example, with Image $\rightarrow$ English and Image $\rightarrow$ German, the generated pseudo (English, German) pair is then pivoted on the Image.
Learning captioning models is practical as it is easier to collect large-scale image-text pairs than translation pairs.
We pre-train these captioning models and use them to generate sentences in two languages depicting the same image, \ie $c^*_x(\mathbf{z}_x)=\text{argmax} p_{z \rightarrow x}(\mathbf{x}|\mathbf{z}_x)$ and $c^*_y(\mathbf{z}_x) = \text{argmax} p_{z \rightarrow y}(\mathbf{y}|\mathbf{z}_x)$. 
The pivoted captions then enable the following two objectives:

\noindent\textbf{Pivoted Captioning for Back-Translation}
We utilize the synthetic multilingual captions (\ie $c^*_x(\mathbf{z}_x)$, $c^*_y(\mathbf{z}_x)$ from the source images and $c^*_x(\mathbf{z}_y)$, $c^*_y(\mathbf{z}_y)$ from the target images) to 
reversely reconstruct the synthetic captions from their translations in both directions.
Formally, we compute the following caption-based back-translation loss:
\begin{equation}
\begin{aligned} \label{cbt_loss}
\mathcal{L}_{x\leftrightarrow y}^{CBT}=
\mathbb{E}_{\mathbf{z}_x}\Big[ 
&\scriptstyle\text{-log } p_{yz\rightarrow x}\big(c^*_x(\mathbf{z}_x)|g^*(c^*_x(\mathbf{z}_x),\mathbf{z}_x), \mathbf{z}_x\big)\\
&\scriptstyle\text{-log } p_{xz\rightarrow y}\big(c^*_y(\mathbf{z}_x)|g^*(c^*_y(\mathbf{z}_x),\mathbf{z}_x), \mathbf{z}_x\big)
\Big]\\
+\mathbb{E}_{\mathbf{z}_y}\Big[
&\scriptstyle\text{-log } p_{yz\rightarrow x}\big(c^*_x(\mathbf{z}_y)|h^*(c^*_x(\mathbf{z}_y),\mathbf{z}_y),\mathbf{z}_y\big) \\
&\scriptstyle\text{-log } p_{xz\rightarrow y}\big(c^*_y(\mathbf{z}_y)|h^*(c^*_y(\mathbf{z}_y),\mathbf{z}_y),\mathbf{z}_y\big)
\Big]
\end{aligned}
\end{equation}

\noindent\textbf{Pivoted Captioning for Paired-Translation}
With the synthetic ``pseudo'' paired (source, target) captions pivoted on a image (\eg $(c^*_y(\mathbf{z}_x)$, $c^*_x(\mathbf{z}_x)$), 
the caption-based paired-translation loss is defined as:
\begin{equation}
\begin{aligned}
\mathcal{L}_{x \leftrightarrow y}^{CPT} &= 
\mathbb{E}_{\mathbf{z}_x}\Big[ \text{-log } p_{xz\rightarrow y}(c^*_y(\mathbf{z}_x)|c^*_x(\mathbf{z}_x), \mathbf{z}_x)\Big]\\
&+
\mathbb{E}_{\mathbf{z}_y}\Big[ \text{-log } p_{yz\rightarrow x}(c^*_x(\mathbf{z}_y)|c^*_y(\mathbf{z}_y), \mathbf{z}_y)\Big]
\end{aligned}
\end{equation}

Note that similar to the text back-translation, for $\mathcal{L}^{CPT}_{x\leftrightarrow y}$ and $\mathcal{L}^{CBT}_{x\leftrightarrow y}$, we do not back-prop through the captioning step.
For optimization, we sample mini-batches and minimizing the following loss: 
\begin{equation}
\mathcal{L} = \mathcal{L}^{MBT}_{x\leftrightarrow y} + \mathcal{L}_{x,y,z}^{VSE} +  \mathcal{L}^{CBT}_{x\leftrightarrow y} +\mathcal{L}^{CPT}_{x\leftrightarrow y}
\end{equation}

Here we drop the weights $w$ of each loss for clarity. 
In practice, all the weights are set to 1.0 except for $w_{CPT}$ where we employ a decreasing learning scheduler specified in the next section.

\section{Experiments and Results}\label{sec_experiment}
We first describe the implementation details and the experimental setup. Then we compare our approach with baselines with detailed analysis.

\subsection{Dataset and Preprocessing} \label{exp_setup}
We conduct experiments on the Multi30K~\cite{multi30k} dataset, the benchmark dataset for multimodal MT.
It contains 29K training, 1K validation, and 1K testing images.
Each image has three descriptions in English/German/French, which are translations of each other.

To ensure the model never learn from parallel sentences, we randomly split Multi30K training and validation sets in half for one language and use the complementary half for the other. The resulting M30k-half are two corpora with non-overlapping 14,500 training and 507 validation image-sentence pairs, respectively.

For text pre-processing, we use Moses~\cite{koehn2007moses} scripts for tokenization and apply the Byte Pair Encoding (BPE)~\cite{sennrich2016neural} from XLM. 
To identify and extract features of visual objects in images, we use the Faster-RCNN~\cite{ren2015faster} model in~\cite{Anderson2017up} to detect up to 36 salient visual objects per image and extract their corresponding 2048-dim regional features.

\subsection{Implementation}
We use Transformer as the underlying architecture for the translation and captioning modules. Each encoder/decoder of the translator is with 6-layer stacked Transformer network, 8 heads, 1024 hidden units, and 4096 feed-forward filter size. 
The captioner is a 6-layer Transformer decoder with the same configuration. 
The visual encoder is a 1-layer MLP which maps visual feature to the shared 1,024-dim embedding space then adds the positional encoding to encode spatial locations (normalized top-left and bottom-right coordinates) of visual objects. 
Our implementation is based on the codebase of XLM and MASS.

\subsection{Experimental Details}
We respectively conduct unsupervised MMT experiments on Multi30K-half for two language pairs: English-French and English-German.

\noindent\textbf{Pre-Training}
Pre-training is a critical step for unsupervised MT. 
We follow the setup in UMMT~\cite{su2019} for a fair comparison. 
For each language, we create a text-only pre-training set by combining the shuffled first 10 million sentences of the WMT News Crawl datasets from 2007 to 2017 with 10 times of M30k-half, resulting in a text-only dataset with 10.145 million unparalleled sentences in English, French, German respectively.

For text pre-training, we leverage the script and the masked seq-to-seq objective proposed in MASS, which randomly masks a span in a sentence then encourages the model to decode and reconstruct the masked sequence as the monolingual language model pre-training. More details can be found in the original paper. Note that there is no fine-tuning (back-translation) on WMT for a fair comparison with other baselines.

For multimodal pre-training of the captioning modules, we use the out-of-domain MS-COCO~\cite{coco} dataset. 
We randomly split the training set into two disjoint subsets. 
Each set contains 56,643 images and 283,215 sentences.
We use the translate-train strategy as in XNLI~\cite{xnli}.
We leverage Google Translate to translate one set of English sentences into French and German.
We pre-train the captioning modules with Eq.~\ref{cap_loss} and fix them during fine-tuning to avoid overfitting.
Note that the captioning modules are trained on non-parallel sentences with disjoint image subsets,
which implies no overlap between English-German or English-French sentences.

\noindent\textbf{Fine-tuning on Multi30K-half} 
We fine-tune on the training set of Multi30K-half for 18 epochs. 
We train our model with the Adam optimizer~\cite{kingma2014adam} with a linear warm-up and a learning rate varying from $10^{-7}$ to $10^{-5}$.
We apply a linearly decreasing weight from 1.0 to 0.1 at 10-th epoch for $w^{CPT}$ as we empirically observe that the generated captions are relatively too noisy to serve as good pseudo pairs in the later stage of training.
The margin $\gamma$ in VSE is set to 0.1. Other hyper-parameters in Transformer follow the default setting in MASS. 
We use 4 Titan Xp GPUs with 1,000 tokens in each mini-batch for training.

\noindent\textbf{Evaluation and Model selection}
For evaluation, we report BLEU scores by multi-bleu.pl\footnote{https://github.com/moses$\text{-}$smt/mosesdecoder/blob/master-/scripts /generic/multi$\text{-}$bleu.perl} in Moses and METEOR\footnote{https://github.com/cmu-mtlab/meteor} scorea on the Multi30K testing set. 

For model selection without a parallel validation corpus, we consider the unsupervised criterion proposed in~\cite{lample2017unsupervised} based on the BLEU scores of ``round-trip'' translations (source $\rightarrow$ target $\rightarrow$ source and target $\rightarrow$ source $\rightarrow$ target) which have been empirically shown to correlate well with the testing metrics.

\begin{table*}[t!]
\centering
\setlength\tabcolsep{2.0pt}
\begin{tabular}{lccccccccc}\hline
 & \multicolumn{2}{c}{en$\rightarrow$fr}  & \multicolumn{2}{c}{fr$\rightarrow$en} & \multicolumn{2}{c}{en$\rightarrow$de} & \multicolumn{2}{c}{de$\rightarrow$en} \\ \hline
Model & \footnotesize BLEU & \footnotesize METEOR & \footnotesize BLEU & \footnotesize METEOR & \footnotesize BLEU & \footnotesize METEOR & \footnotesize BLEU & \footnotesize METEOR \\ \hline 
MUSE\textsuperscript{$\dagger$}~\cite{conneau2017word} & 8.5 & -  & 16.8 & -  & 15.7 & -  & 5.4 & -  \\
UNMT\textsuperscript{$\dagger$}~\cite{lample2017unsupervised} & 32.8 & -  & 32.1 & -  & 22.7 & -  & 26.3 & -  \\
XLM\textsuperscript{$\dagger$}~\cite{xlm}      & 46.3 & 64.3 & 42.0 & 38.1 & 27.4 & 48.7 & 30.7 & 31.0 \\
MASS\textsuperscript{$\dagger$}~\cite{mass}     & 49.8 & 65.8 & 43.7 & 38.7  & 30.2 & 51.3 & 32.5 & 33.4 \\ \hline\hline
Game-MMT~\cite{chen2018zero} & - & - &  - & -  & 16.6 & - & 19.6 & - \\
UMMT-T\textsuperscript{$\dagger$}~\cite{su2019}   & 37.2 & 33.7\textsuperscript{*} & 38.5 & 36.4  & 21.0 & 25.4\textsuperscript{*} & 25.0 & 28.4 \\
UMMT-Full~\cite{su2019} & 39.8 & 35.5\textsuperscript{*} & 40.5 & 37.2 & 23.5 & 26.1\textsuperscript{*} & 26.4 & 29.7 \\ \hline
Ours-Text only\textsuperscript{$\dagger$}   & 49.5 & 65.7 & 43.5 & 38.5 & 30.1 & 51.5 & 32.4 & 33.0 \\
Ours-Full & \textbf{52.3} & \textbf{67.6} & \textbf{46.0} & \textbf{39.8}  & \textbf{33.9} & \textbf{54.1}  & \textbf{36.1} & \textbf{34.7} \\
\hline
\end{tabular}
\caption{\textbf{Results on unsupervised MT}. Comparison with benchmarks on the Multi30K testing set. Our full model is with T+V+VSE+CBT+CPT. 
The best score is marked bold.
\textsuperscript{$\dagger$} means text-only.
\textsuperscript{*} is the METEOR score shown in the UMMT paper.
}\label{result_benchmark}
\vspace{-0.5em}
\end{table*}

\subsection{Baseline Models}
We compare recent unsupervised text-only and multimodal MT baselines listed in the following:
(1) MUSE~\cite{conneau2017word} is a word-to-word MT model with pre-trained Wikipedia embeddings.
(2) UNMT~\cite{lample2017unsupervised} sets the tone of using denoising autoencoder and back-translation for unsupervised MT. 
(3) XLM~\cite{xlm} deploys masked language model from BERT.
(4) MASS~\cite{mass} uses a masked seq-to-seq pre-training objective, achieves the current state-of-the-art performance in text-only unsupervised MT.
(5) Game-MMT~\cite{chen2018zero} is a reinforcement learning-based unsupervised MMT.
(6) UMMT~\cite{su2019} use visual feature for denoising autoencoder and back-translation. 
UMMT is the current state of the art in unsupervised MMT.
We either use the reported scores in the original papers or use their best scripts with their pre-trained language models publicly available for fine-tuning on Multi30K-half.

\subsection{Main Results: Unsupervised MMT}

\subsubsection{Comparison with the Baseline Models}
Table~\ref{result_benchmark} presents the benchmark results with other state-of-the-art unsupervised MT and MMT models on the Multi30K testing set.
The first four rows show the results of the recent text-only MT models.  
Game-MMT and UMMT are MMT models using both image and text inputs.
Our full model (T+V+VSE+CBT+CPT) yields new state-of-the-art performance in BLEU and METEOR, outperforming the text-only and multimodal baseline model by a large margin. 
Notably, our full model outperforms UMMT by +5.5$\sim$12.5 BLEU scores, sets a new state of the art in unsupervised MMT.

Although pre-training plays a vital role in unsupervised MT, comparing Ours-Text only and Ours-Full, the results suggest that multimodal content can further boost the performance for unsupervised MT.
Images provide +2.7$\sim$3.7 BLEU score improvement across four tasks.
Note that our model uses different monolingual pre-training corpora to MASS and XLM for the fair comparison with UMMT.
With a similar pre-training objective, our text-only model is worse than MASS, while Ours-Full outperforms MASS by +2.3$\sim$3.7 in BLEU.

Comparing the multimodal models trained with and without visual content (UMMT-T vs.\ UMMT-Full) and (Ours-T vs.\ Ours-Full), 
our model achieves +2.5$\sim$3.7 improvements in BLEU while +1.4$\sim$2.5 for UMMT.
The results imply that, even with a higher text-only baseline (\eg 49.5 vs.\ 37.2 in en $\rightarrow$fr), pseudo visual pivoting incorporates visual content more effectively.

\begin{figure*}[ht!]
    \centering
    \subcaptionbox{English$\rightarrow$French}[0.498\linewidth]{\includegraphics[width=0.5\linewidth]{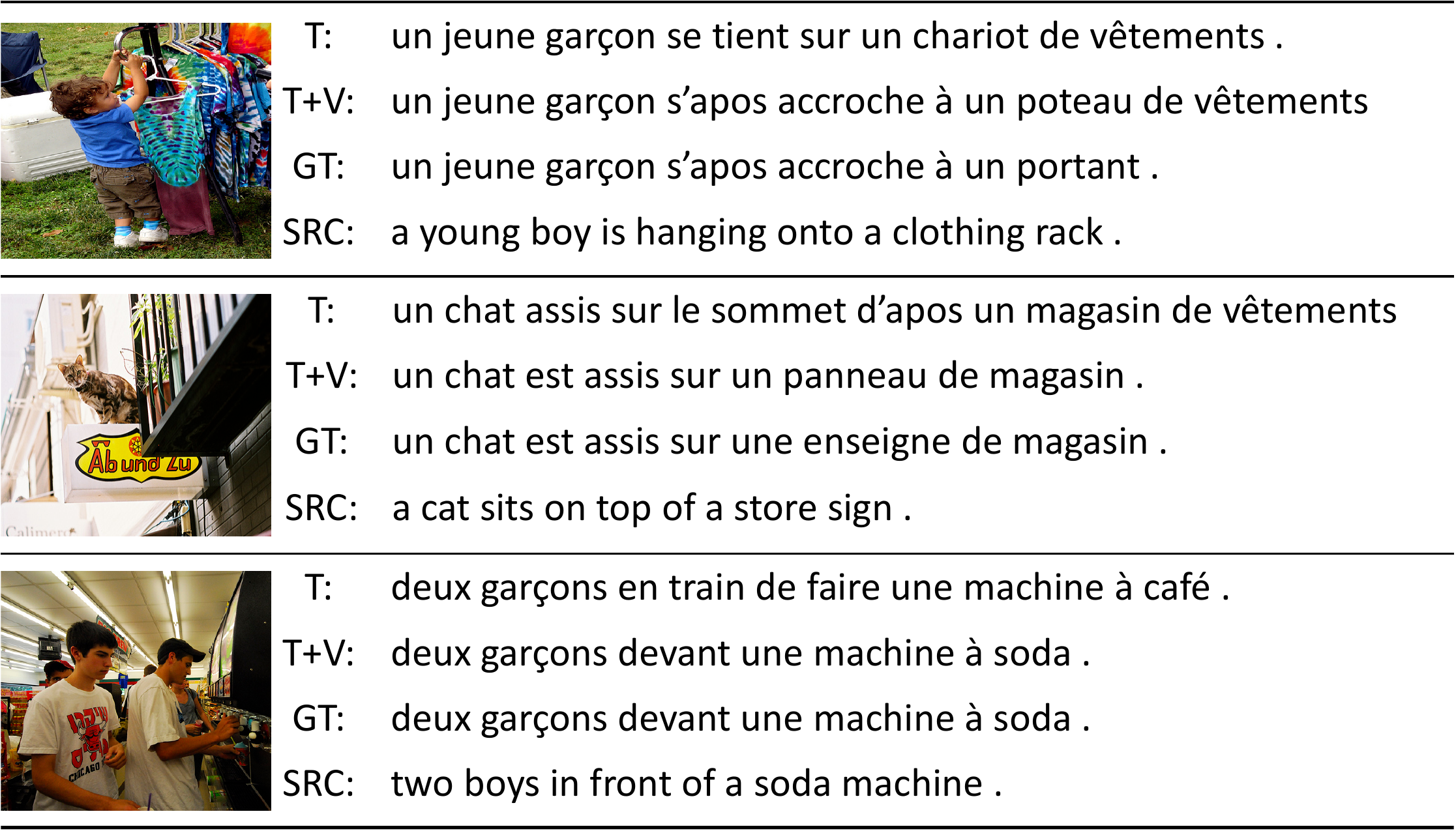}}%
    \subcaptionbox{English$\rightarrow$German}[0.498\linewidth]{\includegraphics[width=0.5\linewidth]{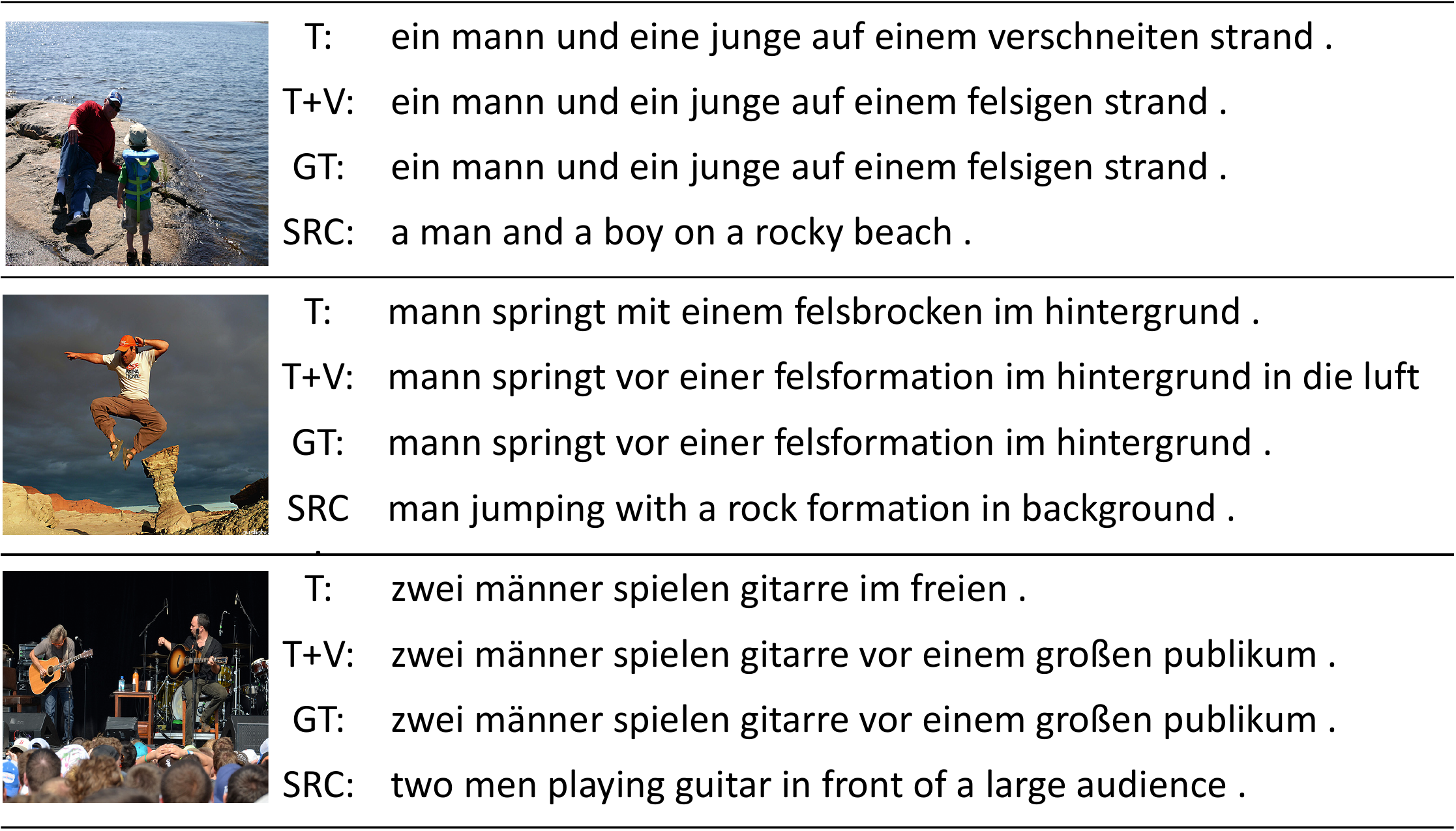}}
    \vspace{-0.5em}
    \caption{Qualitative results of the proposed model. GT: ground truth. T+V: Our full model.}\label{fig_qual}
    \vspace{-0.8em}
\end{figure*}

In Figure~\ref{fig_qual}, we provide some qualitative results on the Multi30K testing set. 
We observe a consistent improvement of unsupervised translation quality with our full model to the text-only one.
Without parallel translation pairs as the vital supervision, the proposed pseudo visual pivoting successfully disambiguates the word semantics in the similar syntactic category and results in improved cross-lingual word alignment;
for instance, ``cafe'' vs.\ ``soda'' machine in the third French example, and ``felsigen'' (rocky) vs.\ ``verschneiten'' (snowy) in the first German example. 

\subsubsection{Ablation Studies}
To quantify module-wise contribution in pseudo visual pivoting, we summarize our ablation studies in Table~\ref{result_ablation}.
Comparing the performance improvement from text-only to the model with regional visual features (T+V),
the features of salient visual objects contribute +0.6$\sim$0.9 BLEU score over a much higher text-only baseline compared to UMMT.

In pseudo visual pivoting,
+VSE promotes the alignments in the monolingual VSE spaces and results in an additional +1.3$\sim$2.0 gain in BLEU.
This improvement validates our hypothesis that the visual space can effectively serve as the bridge connecting the source and target language latent spaces. 
Also, synthesizing image-pivoted pseudo caption pairs effectively provides weak supervision for aligning the cross-lingual latent space in unsupervised MMT.
We observe that the pivoted captions for paired translation (CPT) is more effective than treating them as back-translation pairs (CBT).
Utilizing generated image-pivoted captions is shown to be a promising approach for weakly supervised or unsupervised MMT.
The full model which employs VSE, CBT, and CPT achieves +1.9$\sim$3.1 improvements compared to our multimodal baseline (row two, visual feature only).

\begin{table}[t!]
\centering 
\setlength\tabcolsep{2.5pt}
\begin{tabular}{p{2.6cm}cccc}\hline
Model (Ours) & en$\rightarrow$fr & fr$\rightarrow$en & en$\rightarrow$de & de$\rightarrow$en \\ \hline
Text only & 49.52 & 43.48 & 30.10 & 32.35 \\
T+V & 50.43 & 44.10 & 31.01 & 32.95 \\ 
T+V+VSE & 51.72 & 45.73 & 32.67 & 34.94  \\
T+V+CPT & 51.64 & 45.55 & 33.04 & 35.02 \\
T+V+CBT & 51.23 & 45.21 & 32.51 & 33.87 \\
T+V+VSE+CBT & 51.81 & 45.83 & 33.01 & 34.38 \\
T+V+CPT+CBT & 51.85 & 45.65 & 33.61 & 35.85 \\
T+V+VSE+CPT & 52.19 & \textbf{46.10} & 33.73 & 35.60 \\ \hline
Full Model & \textbf{52.29} & 45.98 & \textbf{33.85} & \textbf{36.07} \\\hline
\end{tabular}
\caption{Ablation studies. BLEU comparison of different training objectives.}\label{result_ablation}
\vspace{-1.0em}
\end{table}

\subsubsection{Generalizability}
How does our unsupervised MMT model generalize when images are not available at the testing time?
Table~\ref{result_generalizability} shows the testing results \textit{without} images.
As can be observed, our model generalizes well. 
The differences are mostly less than 1.0 in BLEU. 
As our model, when being tested without visual content, still outperforms other unsupervised text-only or multimodal MT models listed in Table~\ref{result_benchmark},
the minor drop in BLEU implies that the improved cross-lingual latent space alignment via pseudo visual pivoting is likely to be more critical than using images as an input feature for decoding.
Luckily, such alignment is already preserved in the training phase with the proposed approach.

An interesting question is: How much does the visual content (as a feature) contribute? 
As in leave-one-feature-out cross-validation, we compare the difference of performance between inferencing with and without images.
The larger the difference (the subscripts in Table~\ref{result_generalizability}) implies a model better utilizes visual content.
Compared with UMMT, our model has better utilization. 
We observe that the key to such difference is the VSE objective. 
Our model trained without the VSE objective results in worse utilization (smaller difference at the testing time), possibly because the source text-image pairs are distant in the multilingual VSE space.

\begin{table}[t]
\centering 
\setlength\tabcolsep{1.8pt}
\begin{tabular}{lllll}\hline
Model & en$\rightarrow$fr & fr$\rightarrow$en & en$\rightarrow$de & de$\rightarrow$en \\ \hline
UMMT  & 39.44\tiny-0.35 & 40.30\tiny-0.23 & 23.18\tiny-0.34 & 25.47\tiny-0.92\\\hline
Ours-no VSE  & 51.60\tiny-0.25 & 45.39\tiny-0.26 & 33.25\tiny{-0.36} & \textbf{35.15}\tiny{-0.70} \\
Ours-Full  & \textbf{51.64}\tiny-0.65 & \textbf{45.48}\tiny-0.50 & \textbf{33.32}\tiny{-0.53} & 35.04\tiny{-1.03} \\\hline
\end{tabular}
\caption{BLEU of testing full model with text-only inputs. Subscripts are the difference to testing with T+V.}\label{result_generalizability}
\vspace{-1.0em}
\end{table}

\subsubsection{Real-pivoting \& Low-resource Corpora}
Will our model benefit from ``real'' pivoting (src-img\textsubscript{1}, img\textsubscript{1}-tgt, overall src-img\textsubscript{1}-tgt)?
We train our models with overlapped images while leaving sentences in the source and target languages unparalleled (use \textit{no} translation pairs).
From the first three rows in Table~\ref{result_size}, the performance is improved when training with the overlapped images and their corresponding sentences.
Comparing the improvement from 0\% to 100\% of the text-only model and the full model, a larger gain is observed with the proposed pseudo visual pivoting which aligns and reduces uncertainty in the language latent spaces.

Furthermore, under the low-resource setting (3.0K non-parallel data, row six and seven), a substantial improvement over the text-only model is still observed.
These results suggest that the proposed pseudo visual pivoting is likely to generalize to the semi-supervised and the low-resource setting, which we consider as our future work.

\begin{table}[t]
\centering 
\setlength\tabcolsep{2.7pt}
\begin{tabular}{lcccc}\hline
\makecell{Img overlap \% \\(\# imgs/sents)}&en$\rightarrow$fr&fr$\rightarrow$en& en$\rightarrow$de & de$\rightarrow$en \\ \hline
0\% \footnotesize(14.5K/14.5K) & 52.29 & 45.98 & 33.85 & 36.07 \\
50\% \footnotesize(22K/22K) & 55.13 & 47.54 & 34.61 & 37.01 \\
100\% \footnotesize(29K/29K) & \textbf{58.34} & \textbf{50.57} & \textbf{35.45} & \textbf{38.55} \\\hline
0\% \footnotesize(T only/14.5K) & 49.52 & 43.48 & 30.10 & 32.35 \\
100\% \footnotesize(T only/29K) & 53.35 & 46.27 & 31.35 & 34.06 \\ \hline  \hline
0\% \footnotesize(3.0K/3.0K) & 31.48 & 27.91 & 23.94 & 26.60 \\
0\% \footnotesize(T only/3.0K) & 30.33 & 26.95 & 21.65 & 23.47 \\ \hline
\end{tabular}
\caption{Testing BLEU of the full T+V model and the text-only model trained with overlapped images or low-resource unpaired corpora.}\label{result_size}
\vspace{-1.0em}
\end{table}

\subsubsection{Supervised Case}
Although the proposed pseudo visual pivoting targets unsupervised MMT,
we are also interested in its performance under the fully supervised setup.
To gain insights, we conduct supervised MMT experiments by changing the back-translation objective for unsupervised MT (Eq.~\ref{ummt_loss}) to the supervised MT objective (Eq.~\ref{mt_loss}) with additional visual inputs.
We benchmark with recent supervised MMT models, including Imagination~\cite{imagination}, LIUM-CVC~\cite{caglayan2017lium}, and
VAG~\cite{zhou_visual} on Multi30K.

Table~\ref{result_supervised} shows the testing results. Our model significantly outperforms other baselines and achieves state-of-the-art performance. Comparing to the unsupervised model trained with full Multi30K (Table~\ref{result_size},100\% (29K/29K)), the direct supervision from parallel translation pairs results in a +6.5$\sim$7.1 gain in BLEU. Notably, images provide a minor improvement with full supervision from translation pairs. 
This result implies that, compared to serving as a complementary feature, visual information likely contributes more to improving cross-lingual alignment via pseudo visual pivoting for MMT with limited supervision.

\begin{table}[t]
\centering 
\setlength\tabcolsep{2.6pt}
\begin{tabular}{lcccc}\hline
 & \multicolumn{2}{c}{en$\rightarrow$fr}  & \multicolumn{2}{c}{en$\rightarrow$de} \\\hline
 
Model & \small{BLEU} & \small{METEOR} & \small{BLEU} & \small{METEOR} \\ \hline
Imagination & - & - & 30.2 & 51.2  \\
LIUM-CVC & 52.7 & 69.5 & 30.7 & 52.2  \\
VAG & 53.8 & 70.3 & 31.6 & 52.2  \\
Ours (T) & 65.2 &\textbf{79.3}   & 42.0 & 60.5 \\ 
Ours (T+V)& \textbf{65.5} & 79.1 & \textbf{42.3} & \textbf{60.6}  \\ \hline
\end{tabular}
\caption{Supervised MMT results on Multi30K}\label{result_supervised}
\vspace{-1.0em}
\end{table}

\section{Related Work}
\vspace{-0.5em}

\noindent\textbf{Unsupervised MT} 
For pivoting with a third language,~\newcite{firat2016} pre-train a multi-way multilingual model to generate pseudo pairs to improve zero-shot translation.~\newcite{chen2017} use a teacher-student framework and assume parallel sentences share a similar likelihood for generating sentences in the third language while~\newcite{cheng2017} maximize the expected likelihood. 
Our model does not rely on a third language.
Our framework is along the line of research in \cite{lample2017unsupervised,lample_2018_phrase,xlm}, which aims at learning an aligned latent space between the two languages to translate by reconstruction. 
Nevertheless, we focus on the multimodal setup where the visual space is dissimilar to the language spaces with challenging asymmetric interactions between modalities.

\noindent\textbf{Supervised MMT}
Supervised MMT is introduced in~\cite{mmt_task} as a multi-encoder single-decoder framework with additional image inputs. ~\newcite{mmt} encode word sequences with regional visual objects while ~\newcite{calixto2017incorporating} leverage global visual feature. LIUM-CVC~\cite{caglayan2017lium} uses element-wise multiplication to model the image-text interaction.
Imagination~\cite{imagination} and VAG~\cite{zhou_visual} learns with the auxiliary 
image reconstruction and source-image-target triplet alignment tasks, respectively.
While these methods achieve improvements, their advantage over the text-only models is still minor under the supervised scenario.
As analyzed in~\cite{probing}, visual content is more critical when the textual content is limited or uncertain in MMT. We study the more challenging unsupervised MMT.

\noindent\textbf{Unsupervised MMT} 
To our best knowledge, three recent works have generalized MMT to the unsupervised setting.
~\newcite{nakayama2017zero} learn modal-agnostic fixed length image/sentence embeddings.
In contrast, our model promotes fine-grained (object-token) varying-length embedding, which better aligns VSE space.
Game-MMT~\cite{chen2018zero} use a captioning and a translation model maximizing the likelihood of translated captions to original sentences. 
We synthesize captions for symmetric back-translation and considers no ground truth image annotation in the loop.
Empirically, it is preferred to separate real and generated captions.
UMMT~\cite{su2019} uses Transformers, autoencoder loss, and multimodal back-translation. 
We do not use autoencoder. Our model leverages object detection for multimodal back-translation and equips pseudo visual pivoting.


\noindent\textbf{Image Captioning and VSE} 
Our method draws inspiration from captioning and cross-modal retrieval. 
Recent progress in captioning aims at using reinforcement learning to improve diversity~\cite{dai2017towards} or maximize metric~\cite{self17}.
We use a vanilla MLE objective.
For learning VSE, we leverage the contrastive loss~\cite{KirosSZ14} from cross-modal retrieval, which is shown more robust than maximizing canonical correlation among modalities as in~\cite{andrew2013deep,tweet}.
For encoding image and text, we generalize the cross-modality attention from SCAN~\cite{scan} to the multilingual scenario for learning a multilingual VSE space~\cite{gella2017image, huang_emnlp}.

\vspace{-0.3em}
\section{Conclusion}
\vspace{-0.3em}

We have presented a novel approach: \textit{pseudo visual pivoting} for unsupervised multimodal MT.
Beyond features, we use visual content to improve the cross-lingual alignments in the shared latent space. 
Precisely, our model utilizes the visual space as the approximate pivot for aligning the multilingual multimodal embedding space.
Besides, it synthesizes image-pivoted pseudo sentences in two languages and pairs them to translate by reconstruction without parallel corpora.
The experiments on Multi30K show that the proposed model generalizes well and yields new state-of-the-art performance.

\vspace{-0.3em}
\section*{Acknowledgments}
\vspace{-0.5em}
This work is supported by the DARPA grants funded under the AIDA program (FA8750-18-2-0018), the LWLL program (FA8750-18-2-0501), and the GAILA program (award HR00111990063).
Xiaojun Chang is supported by Australian Research Council Discovery Early Career Award (DE190100626).
The authors would like to thank the anonymous reviewers for their suggestions and Google Cloud for providing the research credits.

\bibliography{acl2020}
\bibliographystyle{acl_natbib}
\end{document}